\def\year{2021}\relax
\documentclass[letterpaper]{article} 
\usepackage{aaai21}  
\usepackage{times}  
\usepackage{helvet} 
\usepackage{courier}  
\usepackage[hyphens]{url}  
\usepackage{graphicx} 
\usepackage{subfigure}
\usepackage{bbm}
\usepackage{multirow}
\usepackage{diagbox} 
\usepackage[switch]{lineno}  %
\usepackage[noend]{algpseudocode}
\usepackage{algorithmicx,algorithm}
\usepackage{cite}
\usepackage{natbib}  
\usepackage{caption} 
\title{Federated Unsupervised Representation Learning}
\author{

    Fengda Zhang\textsuperscript{\rm 1}, Kun Kuang\textsuperscript{\rm 1}, Zhaoyang You\textsuperscript{\rm 1}, Tao Shen\textsuperscript{\rm 1}, Jun Xiao\textsuperscript{\rm 1},\\ Yin Zhang\textsuperscript{\rm 1}, Chao Wu\textsuperscript{\rm 2}\thanks{Corresponding Author}, Yueting Zhuang\textsuperscript{\rm 1}, Xiaolin Li\textsuperscript{\rm 3}
    \\
}
\affiliations{

    \textsuperscript{\rm 1}College of Computer Science and Technology, Zhejiang University\\
    \textsuperscript{\rm 2}School of Public Affairs, Zhejiang University\\
    \textsuperscript{\rm 3}Tongdun Technology

    \{fdzhang, kunkuang, zhaoyangyou, tao.shen\}@zju.edu.cn, junx@cs.zju.edu.cn, \\
    \{zhangyin98, yzhuang, chao.wu\}@zju.edu.cn,  xiaolin.li@tongdun.net

}

\begin{document}

\maketitle

\begin{abstract}
\begin{quote}
To leverage enormous unlabeled data on distributed edge devices, we formulate a new problem in federated learning called \textit{Federated Unsupervised Representation Learning (FURL)} to learn a common representation model without supervision while preserving data privacy.
\textit{FURL} poses two new challenges: 
(1) data distribution shift (Non-IID distribution) among clients would make local models focus on different categories, leading to the inconsistency of representation spaces.
(2) without the unified information among clients in \textit{FURL}, the representations across clients would be misaligned. To address these challenges, we propose Federated Constrastive Averaging with dictionary and alignment (FedCA) 
algorithm. FedCA is composed of two key modules: (1) \textit{dictionary module} to aggregate the representations of samples from each client and share with all clients for consistency of representation space; and (2) \textit{alignment module} 
to align the representation of each client on a base model trained on a public data. We adopt the contrastive loss for local model training. Through extensive experiments with three evaluation protocols in IID and Non-IID settings, we demonstrate that FedCA outperforms all baselines with significant margins.
\end{quote}
\end{abstract}

\section{Introduction}
\textit{Federated Learning (FL)} is proposed as a paradigm that enables distributed clients to collaboratively train a shared model while preserving data privacy \cite{mcmahan2017communication}. Specifically, in each round of federated learning, clients obtain the global model and update it on their own private data to generate the local models, and then the central server aggregates these local models into a new global model. Most of the existing works focus on supervised federated learning in which clients train their local models with supervision. However, the data generated in edge devices are typically unlabeled. Therefore, learning a common representation model for various downstream tasks from decentralized and unlabeled data while keeping private data on devices, i.e. \textit{Federated Unsupervised Representation Learning (FURL)}, remains still an open problem.

\begin{figure}[!t]
\centering                                            
\subfigure[Inconsistency of representation spaces.]{                      
\begin{minipage}{7cm}
\centering                                            
\includegraphics[scale=0.21]{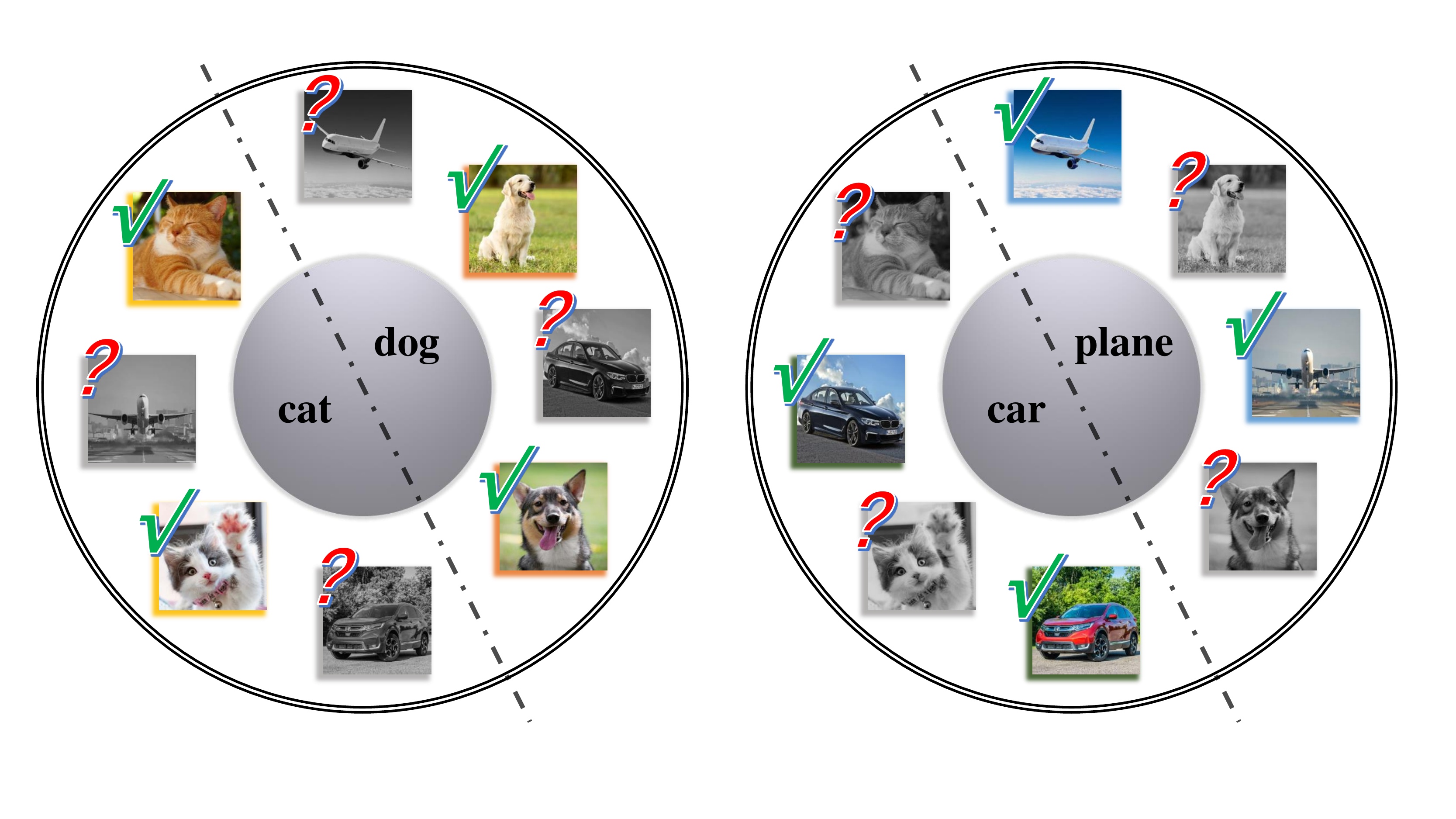}             
\end{minipage}}
\subfigure[Misalignment of representations.]{                   
\begin{minipage}{7cm}
\centering                                            
\includegraphics[scale=0.21]{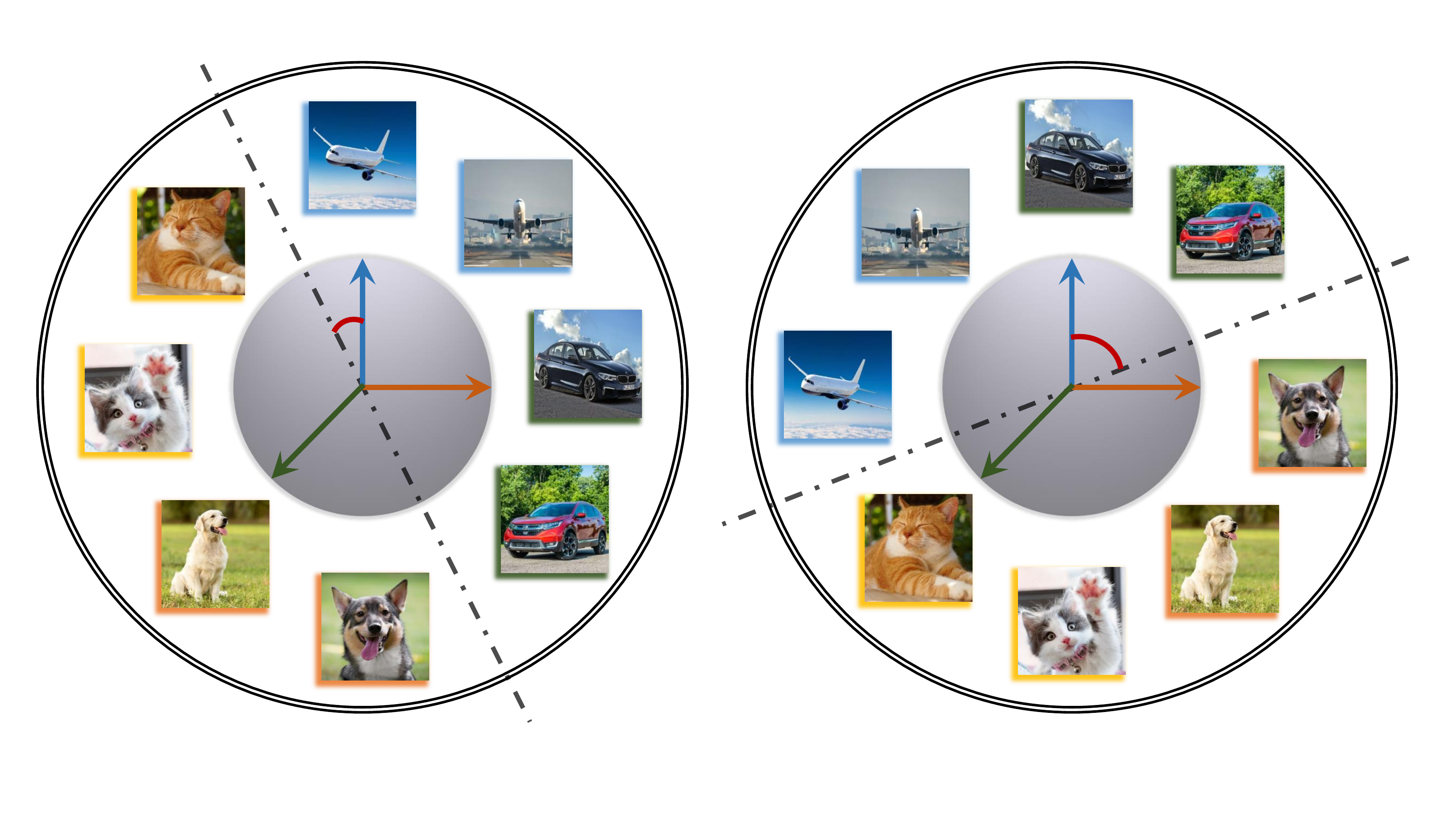}             
\end{minipage}}
\caption{Illustration of challenges in \textit{FURL}: (a) \textit{inconsistency of representation spaces}: data distribution shift among clients causes local models to focus on different categories; and (b) \textit{misalignment of representations}: without unified information, the representation across clients would be misalignment (e.g., rotated by a certain angle). The hyperspheres are representation spaces encoded by different local models in federated learning.}
\label{fig-1}                                         
\end{figure}

It's a natural idea that we can combine federated learning with unsupervised approaches, which means that clients can train their local models via unsupervised methods. There are a lot of highly successful works on unsupervised representation learning. Particularly, contrastive learning methods train models by reducing the distance between representations of positive pairs (e.g., different augmented views of the same image) and increasing the distance between negative pairs (e.g., augmented views from different images), have been outstandingly successful in practice \cite{chen2020simple, oord2018representation, he2020momentum, chen2020improved}. However, their successes highly rely on their abundant data for representation training, for example, contrastive learning methods need a large number of negative samples for training \cite{sohn2016improved, chen2020simple}. Moreover, few of these unsupervised methods take the problem of data distribution shift into account, which is a common practical problem in federated learning. Hence, it's no easy task to combine federated learning with unsupervised approaches for the problem of \textit{FURL}.

In federated learning applications, however, the collected data of each client is limited and the data distribution of client might be different from each other \cite{zhao2018federated, sattler2019robust, jeong2018communication, yang2019federated, kairouz2019advances}. Hence, we face the following challenges to combine federated learning and with unsupervised approaches for \textit{FURL}:
\begin{itemize}
	\item \textbf{Inconsistency of representation spaces.}
	In federated learning, limited data of each client would lead to the variation of data distribution from client to client, resulting in the inconsistency of representation spaces encoded by different local models. For example, as shown in Figure \ref{fig-1}(a), client 1 is with only images of cats and dogs, and client 2 is with only images of cars and planes. Then, the locally trained model on client 1 only encodes a feature space of cats and dogs, failing to map cars or planes to the appropriate representations, and the same goes for trained model on client 2. Intuitively, the performance of the global model aggregated by these inconsistent local models may fall short of expectations.
	\item \textbf{Misalignment of representations.}
	 Even if the training data of clients are IID and the representation spaces encoded by different local models are consistent, there may be misalignment between representations by the reason of randomness in the training process. For instance, for a given input set, the representations generated by a model are equivalent to the representations generated by another model when rotated by a certain angle, as shown in Figure \ref{fig-1}(b). It should be noted that the misalignment between local models may have drastic detrimental effects on the performance of the aggregated model.
\end{itemize}

To address these challenges, we propose a contrastive loss-based federated unsupervised representation learning algorithm called FedCA, which consists of two main novel modules: dictionary module for addressing the inconsistency of representation spaces and alignment module for aligning the representations across clients.
Specifically, the dictionary module, which is maintained by server, aggregates abundant representations of samples from clients and can be shared to each client for local model optimization. 
In the alignment module, we first train a base model based on a small public data (e.g., a subset of STL-10 dataset) \cite{coates2011analysis}, then require all local models to mimic the base model such that the representations generated by different local models can be aligned. 
Overall, in each round, FedCA involves two stages: (i) \textit{clients} train local representation models on their own unlabeled data via contrastive learning with two modules above, and then generate local dictionaries, and (ii) \textit{server} aggregates the trained local models to obtain a shared global model and integrates local dictionaries into a global dictionary.

To the best of our knowledge, FedCA is the first algorithm designed for the \textit{FURL} problem. 
Our experiments show that FedCA has better performance than those naive methods which solely combine federated learning with unsupervised approaches. We believe that FedCA will serve as a critical foundation in this novel and challenging problem.

\begin{figure*}[!t]
\centering                                            
\subfigure[Overview of FedCA.]{                
\begin{minipage}[t]{0.27\linewidth}
\centering                                            
\includegraphics[width=1.9in]{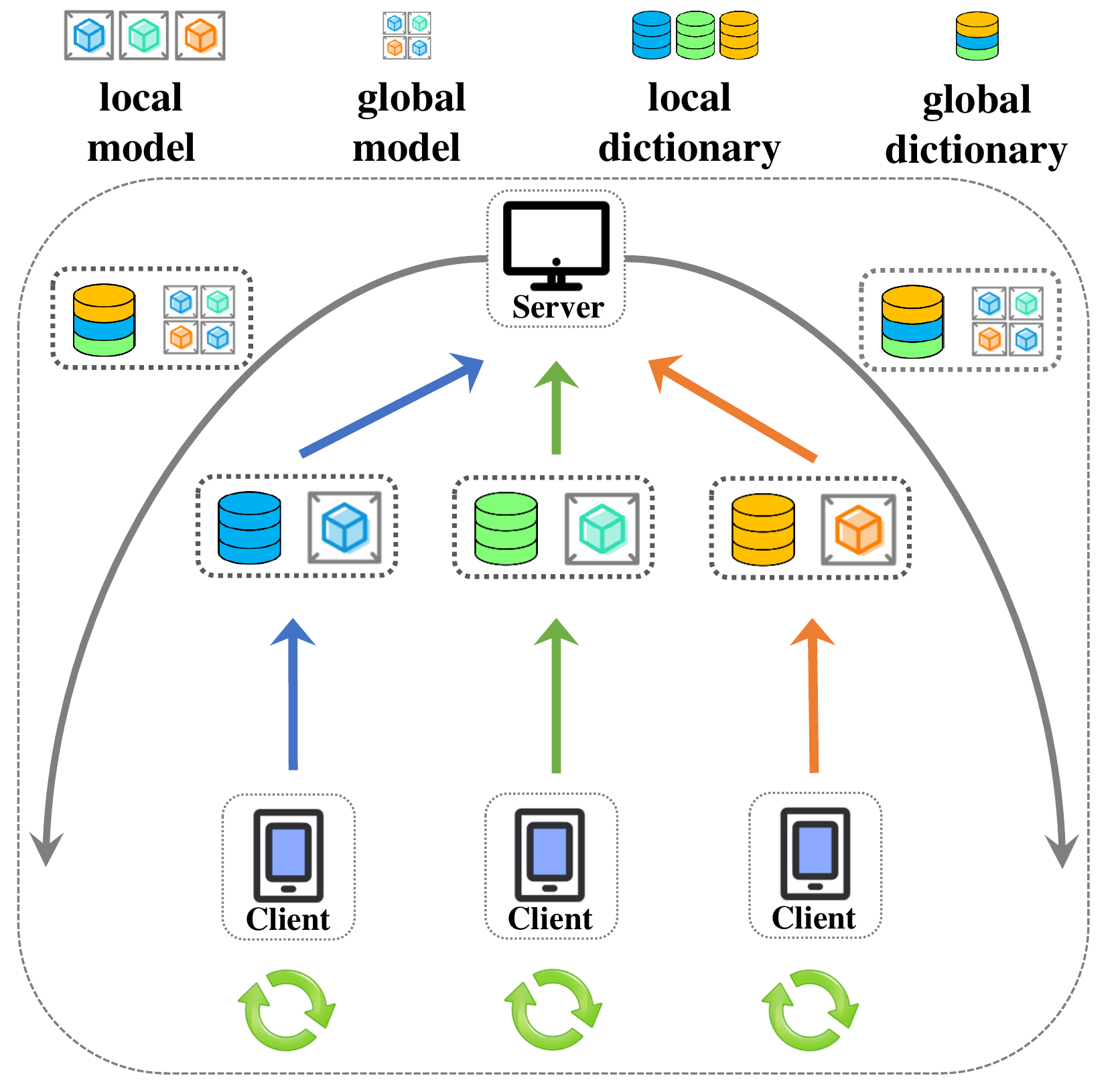}             
\end{minipage}}
\subfigure[Local Update of Model.]{                
\begin{minipage}[t]{0.45\linewidth}
\centering                                            
\includegraphics[width=2.7in]{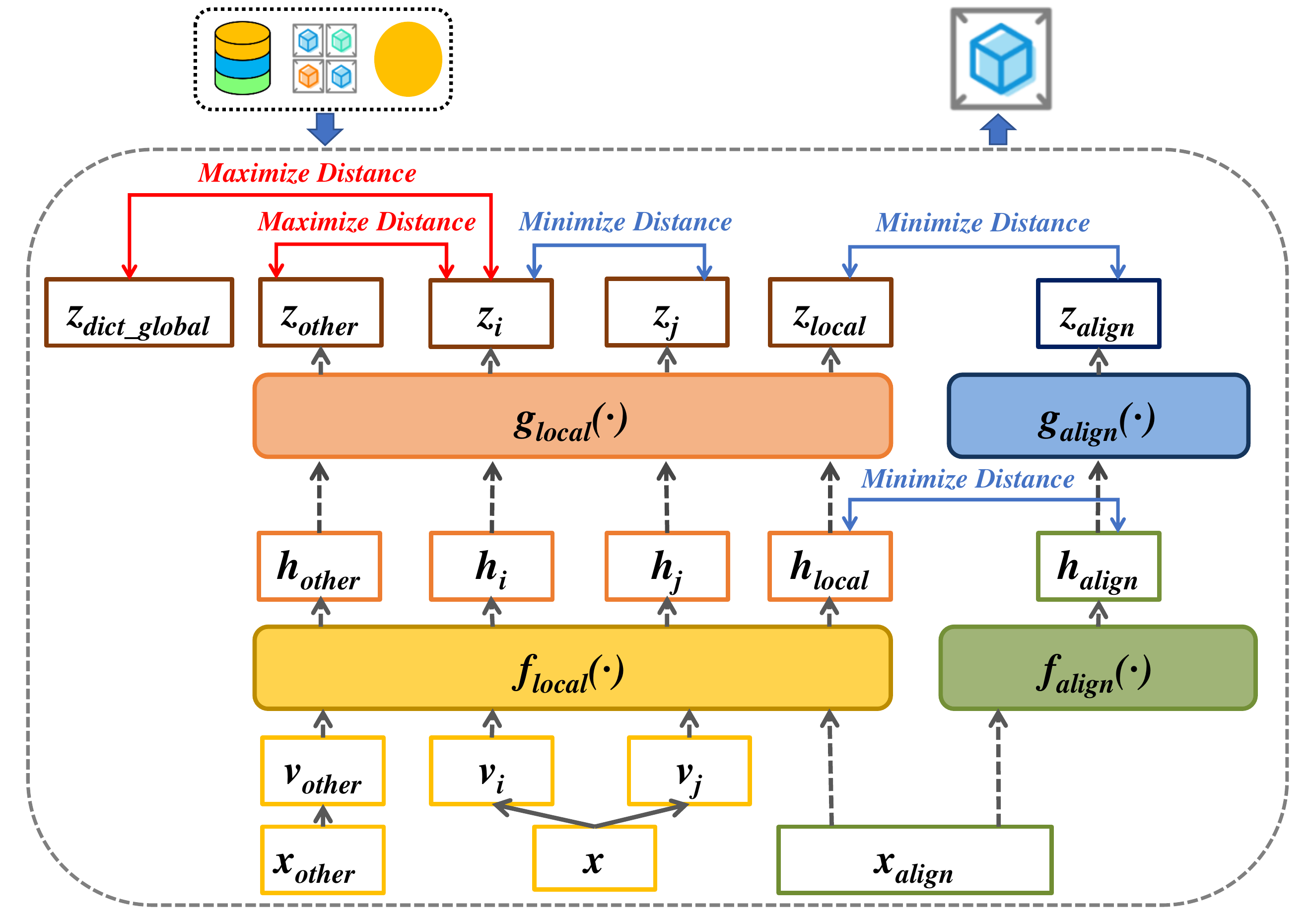}             
\end{minipage}}
\subfigure[Local Update of Dictionary.]{                
\begin{minipage}[t]{0.23\linewidth}
\centering                                            
\includegraphics[width=1.7in]{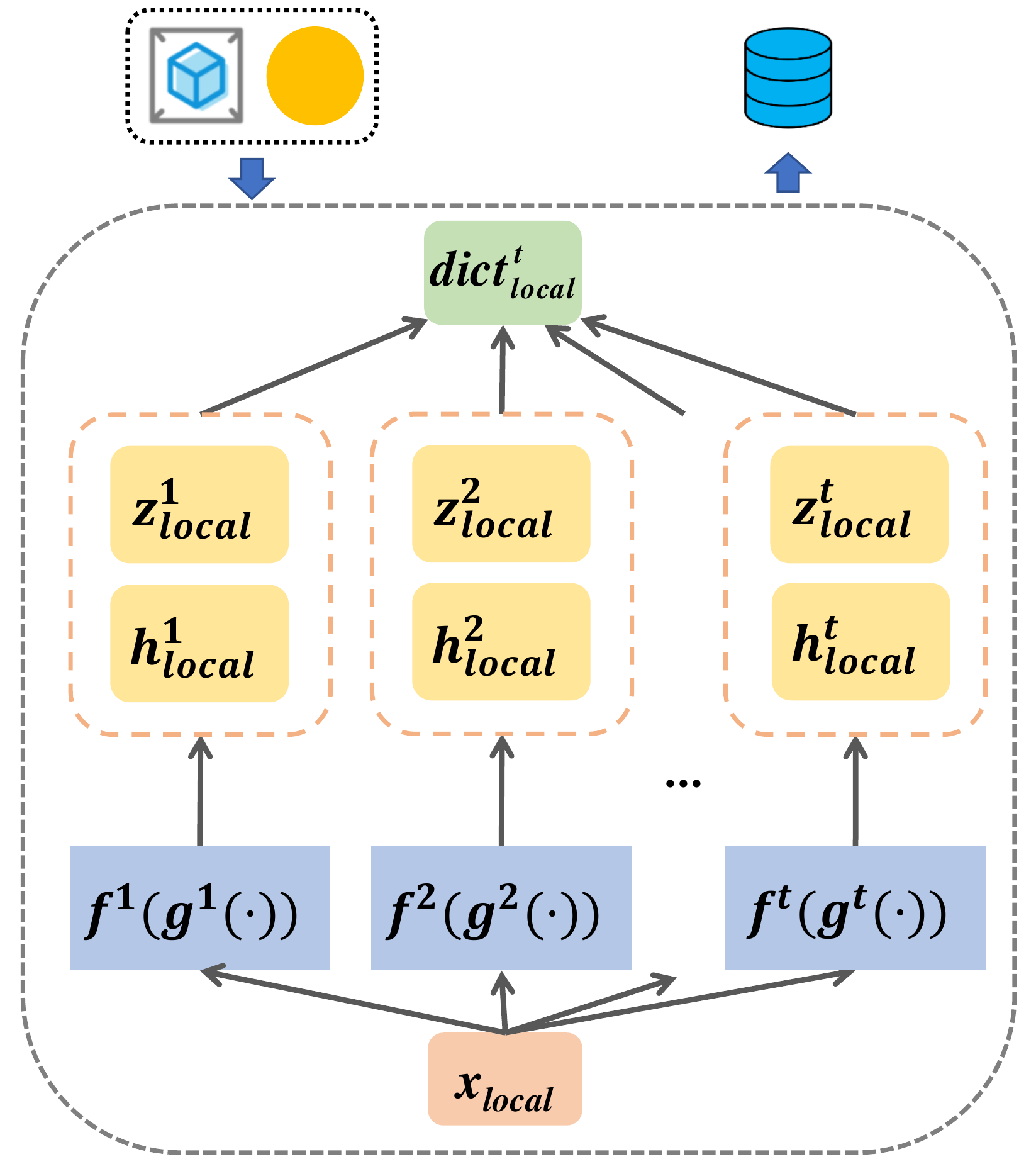}             
\end{minipage}}
\caption{Illustrations of FedCA. (a) In each round, clients generate local models and dictionaries, and then server gathers them to obtain global model and dictionary. (b) Clients update local models by contrastive leaning with the dictionary and alignment modules. $x_{other}$ is a sample different from sample $x$, and $x_{alignment}$ is a sample from the additional public dataset for alignment. $f$ is the encoder and $g$ is the projection head. (c) Clients generate local dictionaries via temporal ensembling.}                   
\label{fig-2}                                         
\end{figure*}

~\\
\section{Related Work}
\subsection{Federated Learning}
Federated learning enables distributed clients to train a shared model collaboratively while keep private data on devices \cite{mcmahan2017communication}. Li et al. add a proximal term to the loss function to keep local models close to the global model \cite{li2018federated}. Wang et al. propose a layers-wise federated learning algorithm to deal with permutation invariance of neural network parameters \cite{wang2020federated}. However, the existing works only focus on the consistency of parameters, while we emphasize the consistency of representations in this paper. Some works also focus on reducing the communication of federated learning \cite{konevcny2016federated}.
To further protect data privacy of clients, cryptography technologies are applied to federated learning \cite{bonawitz2017practical}.
\subsection{Unsupervised Representation Learning}
There are two main types of unsupervised learning methods: generative and discriminative. 
Generative approaches learn representations by generating pixels in the input space \cite{hinton2006reducing, kingma2013auto, radford2015unsupervised}. 
Discriminative approaches train representation model by performing pretext tasks where the labels are generated for free from unlabelled data \cite{pathak2017curiosity, gidaris2018unsupervised}.
Among them, contrastive learning methods achieve excellent performance \cite{chen2020simple, oord2018representation, he2020momentum, chen2020improved}. The contrastive loss is proposed by Hadsell et al. \cite{hadsell2006dimensionality}. Wu et al. propose an unsupervised contrastive learning approach based on a memory bank to learn visual representations \cite{wu2018unsupervised}. Recently, Wang et al. point two key properties, closeness and uniformity, related to the contrastive loss. \cite{wang2020understanding}.
Other works also apply contrastive learning to video \cite{sermanet2018time, tian2019contrastive}, NLP \cite{mikolov2013distributed, logeswaran2018efficient, yang2019xlnet}, audio \cite{baevski2020wav2vec} and graph \cite{hassani2020contrastive, qiu2020gcc}.
\subsection{Federated Unsupervised Learning}
Some concurrent works \cite{Jin2020TowardsUU, van2020towards} also focus on federated learning from unlabeled data. Different from these works that all simply combine federated learning with unsupervised approaches, we explore and identify the main challenges in federated unsupervised representation learning and design an algorithm to deal with these challenges.

~\\
\section{Preliminary}
In this section, we discuss the primitives needed for our approach.
\subsection{Federated Learning}
In federated learning, each client $u \in U$ has a private dataset $D_u$ of training samples with $D = \cup_{u \in U} D_u $ and our aim is to train a shared model while keeping private data on devices. There are a lot of algorithms designed for aggregation in federated learning \cite{wang2020federated, li2018federated}. Here, for simplicity, we introduce a standard and popular aggregation method named FedAvg \cite{mcmahan2017communication}. In each round of FedAvg, the server randomly selects a subset of clients $U^t \subseteq U$ and each client $u \in U^t$ locally updates the model $f$ with parameters $\theta_t$ on dataset $D_u$ via stochastic gradient descent rule:
\begin{equation}
\theta^{t+1}_u \gets \theta^t - \eta\bigtriangledown\mathcal{L}_f(D_u, \theta^t)
\end{equation}
where $\eta$ is the stepsize. Then the server gathers parameters of local models $\{\theta^{t+1}_u | u \in U^t\}$ and aggregate these local models via weighted average to generate a new global model:
\begin{equation}
\theta^{t+1} \gets \sum_{u \in U^t} \frac{|D_u|}{\sum_{i \in U^t}|D_i| } \theta^{t+1}_u
\end{equation}
The training process above is repeated until the global model converges.

\subsection{Unsupervised Contrastive Learning}
Unsupervised contrastive representation learning methods learn representations from unlabeled data by reducing the distance between representations of positive samples and increasing the distance between representations of negative samples.
Among them, SimCLR achieves outstanding performance and can be applied to federated learning easily \cite{chen2020simple}. SimCLR randomly samples a minibatch of $N$ samples and executes twice random data augmentations for each sample to obtain $2N$ views. Typically, the views augmented from the same image are treated as positive samples and the views augmented from different images are treated as negative samples \cite{dosovitskiy2014discriminative}. The loss function for a positive pair of samples $(i, j)$ is defined as:
\begin{equation}
l_{i, j} = -log\frac{exp(sim(z_i, z_j)/\tau)}{\sum_{k=1}^{2N}\mathbbm{1}_{[k \neq i]}exp(sim(z_i, z_k)/\tau)},
\end{equation}
where $\tau$ is temperature and $\mathbbm{1}_{[k \neq i]} = 1$ iff $k \neq i$.
The model (consisting of a base encoder network $f$ to extract representation vectors $h$ from augmented views and a projection head $g$ to map representations $h$ to $z$) is trained by minimizing the loss function above. Finally, we use representations $h$ to perform downstream tasks.

~\\
\section{Method}
In this section, we analyze two challenges mentioned above and detail the dictionary module and alignment module designed for these challenges. Then we introduce Federated Contrastive Averaging with dictionary and alignment (FedCA) algorithm for \textit{FURL}.
\subsection{Dictionary Module for Inconsistency Challenge}
\textit{FURL} aims to learn a shared model mapping data to representation vectors such that similar samples are mapped to nearby points in representation space so that the features are well-clustered by classes.
However, the presence of Non-IID data presents a great challenge to \textit{FURL}. Since the local dataset $D_u$ of a given client $u$ likely contains samples of only a few classes, the local models may encode inconsistent spaces, causing bad effects on the performance of the aggregated model.

\begin{figure}[!t]
\centering                                            
\subfigure[Vanilla Federated Unsupervised Approach.]{
\begin{minipage}{7cm}               
\includegraphics[scale=0.20]{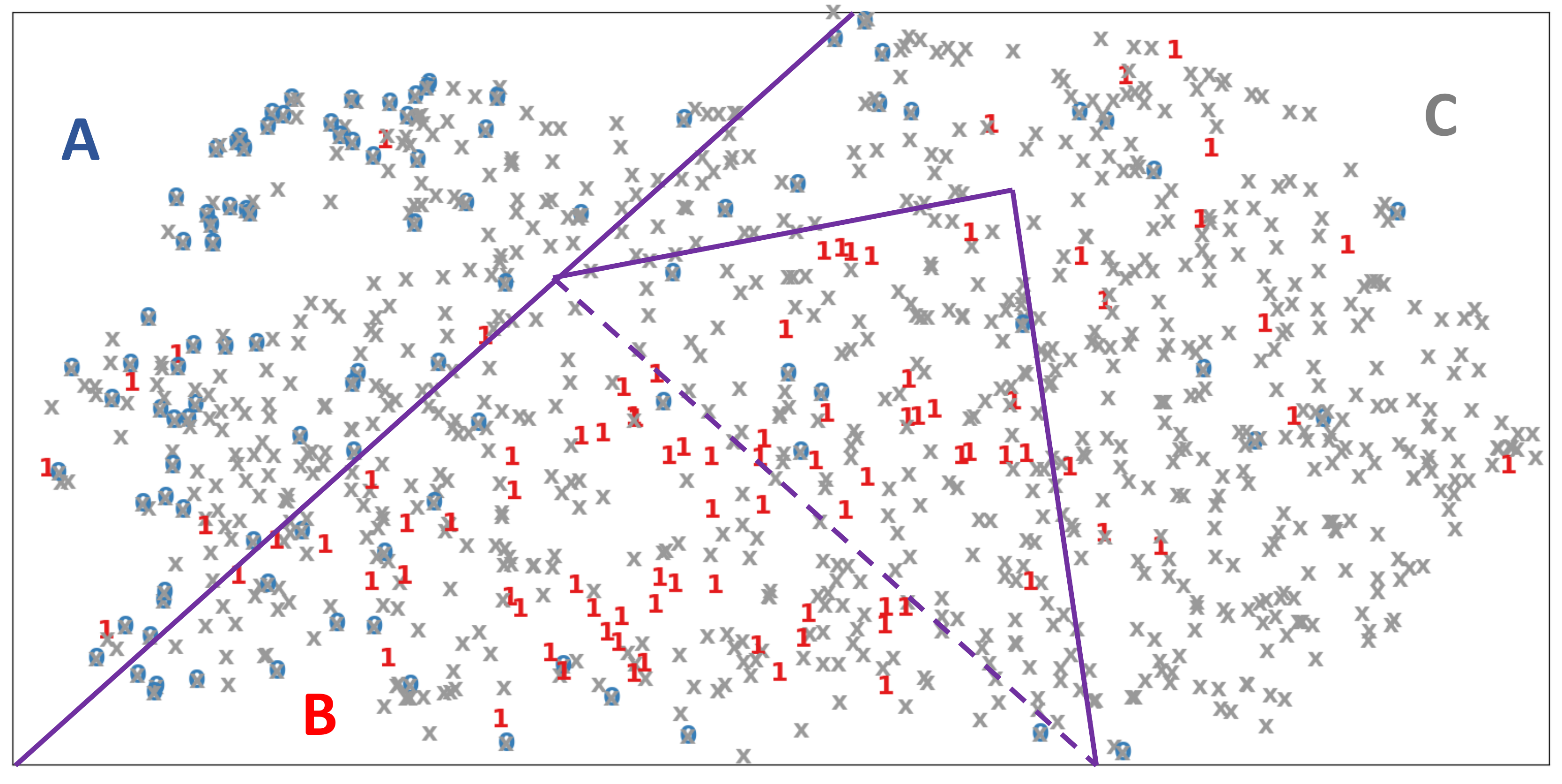}             
\end{minipage}}
\subfigure[FedCA.]{
\begin{minipage}{7cm}
\includegraphics[scale=0.20]{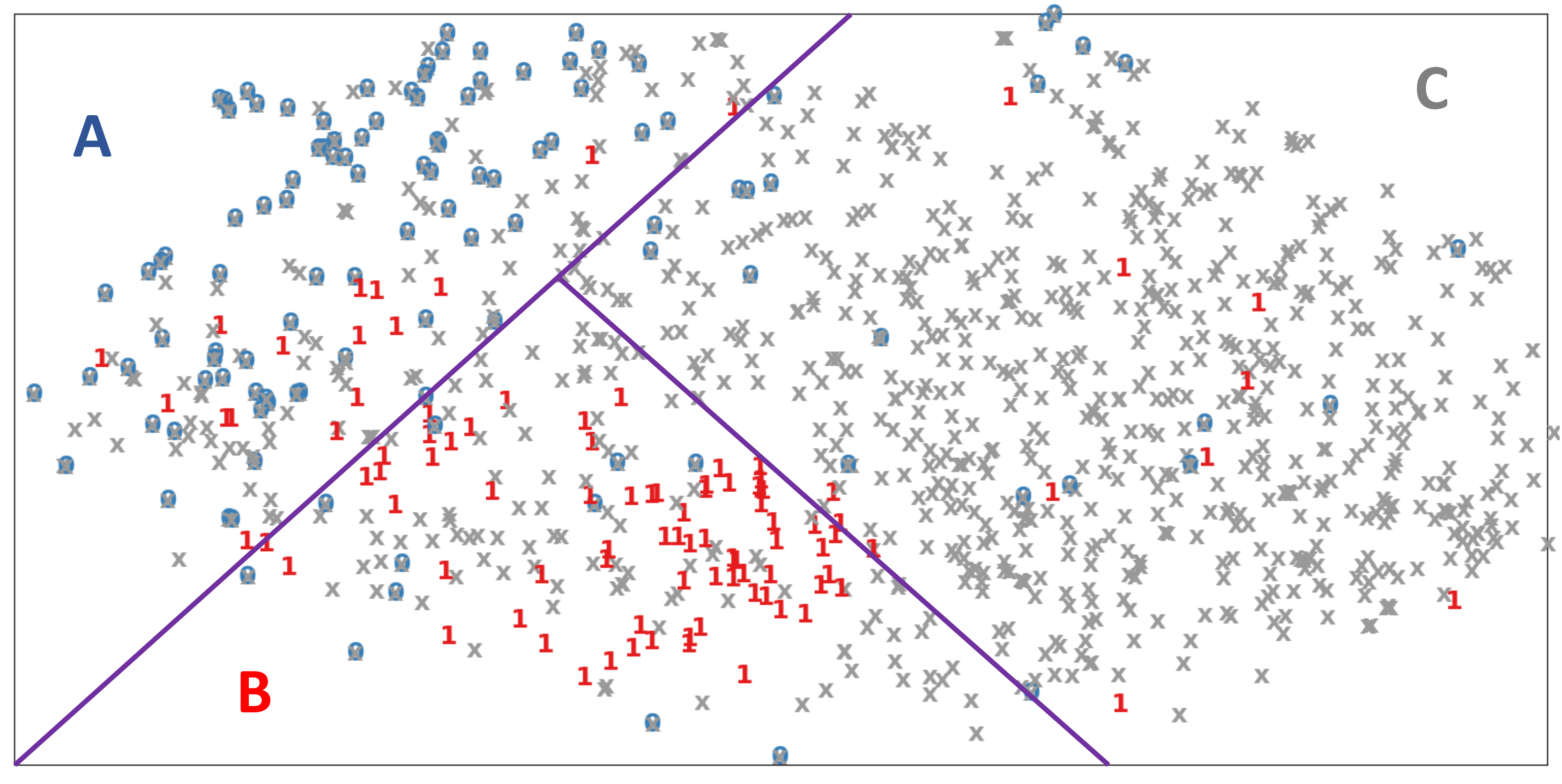}             
\end{minipage}}
\caption{T-SNE visualization results of representations on CIFAR10. In federated learning with Non-IID setting, we use the local model of the client who only has samples of class 0 and 1 to generate representations. We compare two methods: (a) FedSimCLR (SimCLR is combined with FedAvg directly) and (b) FedCA (ours). A, B are the regions where representations of samples of class 0, 1 cluster respectively and C is the rest region.} 
\label{fig-3}                                         
\end{figure}

To empirically verify this, we visualize the representations of images from CIFAR-10 via T-SNE method. To be specific, we split training data of CIFAR-10 into 5 Non-IID sets and each set consists of 10000 samples from 2 classes. Then FedAvg algorithm is solely combined with unsupervised approach (SimCLR) to learn representations from these subsets. We use the local model in 20th round of the client who only has samples of class 0 and 1 to extract features from test set of CIFAR-10 and visualize the representations after dimensionality reduction by T-SNE, as shown in Figure \ref{fig-3}(a). We find that the scattered representations of samples from class 0 and 1 spread over a very large area of representation space and it is difficult to distinguish samples of class 0 and 1 from others. It suggests that the local model encodes a representation space of samples of class 0 and 1 and it cannot map samples of other classes to the suitable positions. The visualization results support our hypothesis that the representation spaces encoded by different local models are inconsistent in Non-IID setting.

We argue that the cause of the inconsistency is that the clients can only use their own data to train the local models but the distribution of data varies from client to client. To address this issue, we design a dictionary module, as shown in Figure 2(b). Specifically, in each communication round, clients use the global model (including the encoder and the projection head) to obtain the normalized projections $\{\tilde{z_i}\}$ of their own samples and send normalized projections to the server along with the trained local models. Then the sever gathers the normalized projections into a shared dictionary. For each client, the global dictionary $\tilde{z}_{dict}$ with $K$ projections is treated as a normalized projection set of negative samples for local contrastive learning.
Specifically, in local training process, for a given minibatch $x_{batch}$ with $N$ samples, we randomly augment them to obtain $x_i$, $x_j$ and generate normalized projections $\tilde{z_i}$, $\tilde{z_j}$. Then we calculate
\begin{equation}
logits_{batch} = \tilde{z_i} \cdot {\tilde{z_j}}^T,
\end{equation}
\begin{equation}
logits_{dict} = \tilde{z_i} \cdot {\tilde{z}_{dict}}^T,
\end{equation}
\begin{equation}
logits_{total} = concat([logits_{batch}, logits_{dict}], dim=1),
\end{equation}
where $concat()$ denotes concatenation and the size of $logits$ is $N \times (N+K)$. Now we turn the unsupervised problem into a $(N+K)$-classification problem and define
\begin{equation}
label = [0, 1, 2, ...,N-2, N-1]
\end{equation}
as a class indicator. Then the loss function is given as \begin{equation}
loss_{contrastive} = CE(logits/t, labels),
\end{equation}
where $CE()$ denotes cross entropy loss and $t$ is temperature term.

Note that, in each round, the shared dictionary is generated by global model from the previous round, but the projections of local samples are encoded by current local models. The inconsistencies in representations may affect the function of the dictionary module, especially in Non-IID setting. We use temporal ensembling to alleviate this problem, as shown in Figure \ref{fig-2}(c). To be specific, each client maintains a local ensemble dictionary consisting of projections set $\{Z_i^{t-1}| x_i \in D_u\}$. In each round, client $u$ uses trained local model to obtain projections $\{z_i^{t}| x_i \in D_u\}$ and accumulates it into ensemble dictionary by updating 
\begin{equation}
Z_i^t \leftarrow \alpha Z_i^{t-1} + (1-\alpha) z_i^t, 
\end{equation}
and then normalized ensemble projection is given as
\begin{equation}
\tilde{z_i^t} = \frac{Z_i^t / (1-{\alpha}^t)}{||Z_i^t / (1-{\alpha}^t)||_2} = \frac{Z_i^t}{||Z_i^t||_2},
\end{equation}
where $\alpha \in [0, 1)$ is a momentum parameter and $Z_i^0 = \vec{0}$.

We visualize the representations encoded by local model trained via federated contrastive learning with dictionary module in the same setting as vanilla federated unsupervised approach. As shown in Figure \ref{fig-3}(b), we find that the points of class 0 and 1 are clustered in a small subspace of representation space, which means that the dictionary module works well as we expected.

\subsection{Alignment Module for Misalignment Challenge}

\begin{figure}[t]
\centering                                            
\subfigure[FedSimCLR.]{
\begin{minipage}[t]{0.48\linewidth}
\includegraphics[width=1.55in]{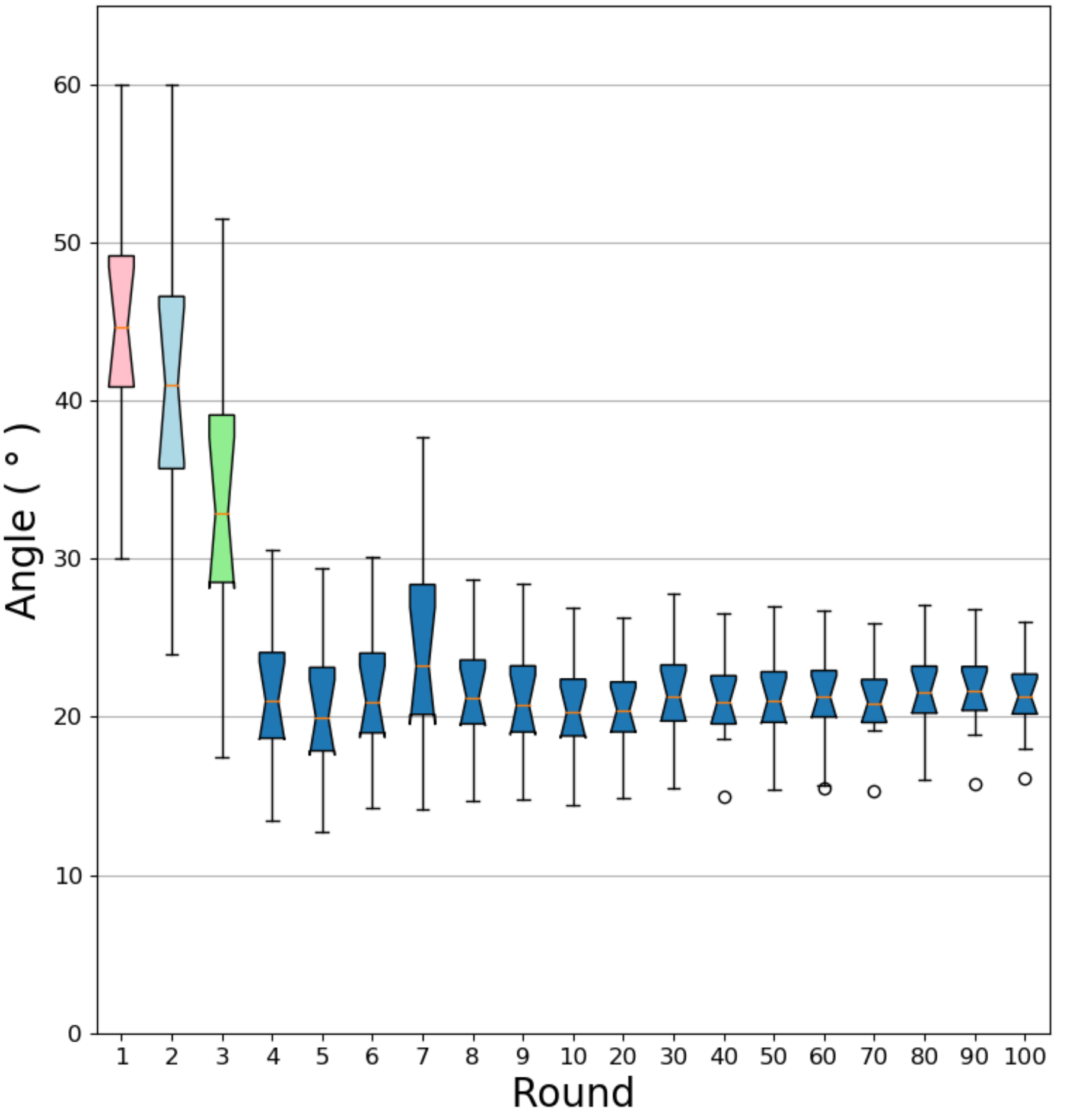}             
\end{minipage}}
\subfigure[FedCA.]{
\begin{minipage}[t]{0.48\linewidth}
\includegraphics[width=1.55in]{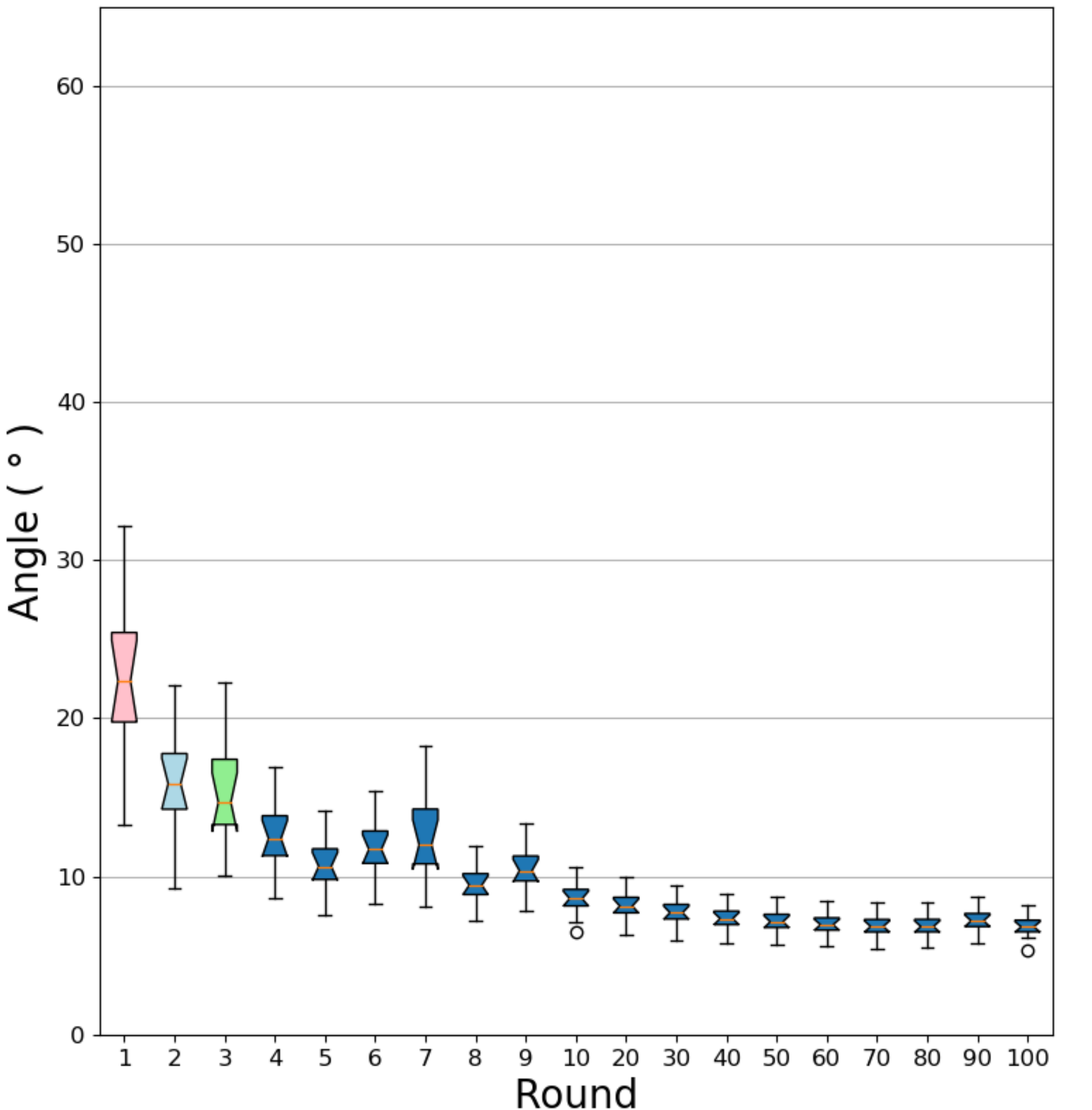}             
\end{minipage}}
\caption{Boxplots of angles between representations encoded by local models on CIFAR10 in federated learning with IID setting.} 
\label{fig-4}                        
\end{figure}

Due to the randomness in training process, there might be a certain angle difference between representations generated by two models trained on the same dataset respectively, although these two models encode consistent spaces. The \textit{misalignment of representations} may have an adverse effect on model aggregation.

To verify it, we record the angles between normalized representations generated by different local models in federated learning. We split training data of CIFAR-10 into 5 IID sets randomly and each set consists of 10000 samples from all 10 classes. We randomly select 2 local models trained by vanilla federated unsupervised approach (FedSimCLR is used as an example) and use them to obtain normalized representations on testset of CIFAR-10. As shown in Figure \ref{fig-4}(a), there is always a large angle (beyond $20^{\circ}$) difference between representations encoded by the local models in learning process.

We introduce an alignment module to tackle this challenge. As shown in Figure \ref{fig-2}(b), we prepare an additional public dataset $D_{align}$ with small size and train a model $g_{align}(f_{align}())$ (called alignment model) on it. The local models are then trained via contrastive loss with a regularization term that replicating outputs of the alignment model on a subset of alignment dataset. For a given client $u$, the loss function is defined as
\begin{equation}
loss_{align}^h = \sum_{i=1}^{|D_{align}^{sub}|}||h_{align}^i - h_{u}^i||_2^2,
\end{equation}
\begin{equation}
loss_{align}^z = \sum_{i=1}^{|D_{align}^{sub}|}||z_{align}^i - z_{u}^i||_2^2,
\end{equation}
\begin{equation}
loss_{align} = loss_{align}^z + loss_{align}^z,
\end{equation}
where $h_{align}^i = f_{align}(x^i)$, $z_{align}^i = g_{align}(h_{align}^i)$, $h_u^i = f_u(x^i)$, $z_u^i = g_u(h_u^i)$, $x^i \in D_{align}^{sub} \subseteq D_{align}$.

We also calculate the angles between representations of local models trained via federated contrastive learning with alignment module (3200 images sampled from STL-10 randomly are used for alignment) in the same setting as vanilla federated unsupervised approach. As shown in Figure \ref{fig-4}(b), the angles can be controlled within $10^{\circ}$ after 10 training rounds, which suggests that the alignment module can help to align the local models.

\subsection{FedCA Algorithm}

From the above, the total loss function of local model update is given as
\begin{equation}
     loss = loss_{contrastive} + \beta loss_{align},
\end{equation}
where $\beta$ is a scale factor controlling the influence of alignment module.
Now we have a complete algorithm named Federated Contrastive Averaging with Dictionary and Alignment (FedCA) which can handle the challenges of \textit{FURL} well, as shown in Figure \ref{fig-2}.

\begin{algorithm}[!h]
\caption{\textit{Federated Contrastive Averaging with Dictionary and Alignment (FedCA)}.} 
{\textbf{Require:} The $n$ clients are indexed by $u$; parameters of global model (encoder and projection head) $\theta_t$, parameters of local model $\theta_t^u$, global dictionary $dict_t$, local dictionary $dict_t^u$, the proportion of selected clients $C$, the number of local epochs $E$, local dataset $D_u$, and learning rate $\eta$.} 

\hspace*{0.02in} {\textbf{Server executes:}} 
\begin{algorithmic}[1]
\State Initialize $\theta_0$
\State Prepare a public dataset $D_{align}$ and an alignment model with parameters $\theta_{align}$
\For{each round $t = 0, 1, 2, ...$}
    \State $m \leftarrow max(C \cdot n, 1)$
    \State $U_t \leftarrow$ (random set of $m$ clients)
    \For{each client $u \in U_t$ \textbf{in parallel}}
        \State $\theta_{t+1}^u, dict_{t+1}^u \leftarrow ClientUpdate(u, \theta_t, dict_t)$ 
    \EndFor
    \State $\theta_{t+1} \leftarrow \sum_{u \in U_t} \frac{|D_u|}{\sum_{i \in U_t}|D_i|} \theta_{t+1}^u$ 
    \State $dict_{t+1} \leftarrow concat([\{dict_{t+1}^{u}|u \in U_t\}], dim=1)$
\EndFor
\end{algorithmic}
\hspace*{0.02in} {\textbf{ClientUpdate}($u$, $\theta$, $dict$) \textbf{:}} 
// Run on client $u$
\begin{algorithmic}[1]
\For{each local epoch $i$ from $1$ to $E$}
    \For{batch $b \in D_u$}
        \State // Update $\theta$ with Eq. (14)
        \State $\theta \leftarrow \theta - \eta \bigtriangledown\mathcal{L}(\theta; b, dict, D_{align}, \theta_{align})$
    \EndFor
\EndFor
\State Generate $dict^u$ by Eq. (9)(10)
\State \Return $\theta$, $dict^u$
\end{algorithmic}
\end{algorithm}

Algorithm 1 summarizes the proposed approach.

\section{Experiments}
\textit{FURL} aims to learn a representation model from decentralized and unlabeled data. In this section, we present an empirical study of FedCA.

\begin{table*}[t]
  \centering
  \fontsize{6.5}{8}\selectfont
  \label{tab:performance_comparison}
    \begin{tabular}{|c|c|c|c|c|c|c|c|}
    \hline
    \multirow{2}{*}{Setting}&
    \multirow{2}{*}{Method}&
    \multicolumn{2}{c|}{CIFAR10}&\multicolumn{2}{c|}{CIFAR100}&\multicolumn{2}{c|}{MiniImageNet}\cr\cline{3-8}
    & &5-layer CNN&ResNet-50&5-layer CNN&ResNet-50&5-layer CNN&ResNet-50\cr
    \hline
    \multirow{4}{*}{IID}&
    FedAE&61.23&65.47&34.07&36.56&28.21&31.97\cr
    &FedPR&55.75&63.52&29.74&30.89&24.76&26.63\cr
    &FedSimCLR&61.62&68.10&34.18&39.75&29.84&32.18\cr
    &FedCA (ours)&{\bf 64.87}&{\bf 71.25}&{\bf 39.47}&{\bf 43.30}&{\bf 35.27}&{\bf 37.12}\cr\hline
    \multirow{4}{*}{Non-IID}&
    FedAE&60.14&63.74&33.94&37.27&29.00&30.44\cr
    &FedPR&54.94&60.31&30.70&32.39&24.74&25.91\cr
    &FedSimCLR&59.21&64.06&33.63&38.70&29.24&30.47\cr
    &FedCA (ours)&{\bf 63.02}&{\bf 68.01}&{\bf 38.94}&{\bf 42.34}&{\bf 34.95}&{\bf 35.01}\cr
    \hline
    \end{tabular}
  \caption{Top-1 accuracies (\%) of algorithms for \textit{FURL} on linear evaluation}
\label{table-1} 
\end{table*}

\subsection{Experimental Setup}
\subsubsection{Baselines.} \textit{AutoEncoder} is a generative method to learn representations in an unsupervised manner by generating from the reduced encoding a representation as close as possible to its original input \cite{hinton2006reducing}. \textit{Predicting Rotation} is one of the proxy tasks of self-supervised learning by rotating samples by random multiples of 90 degrees and predicting the degrees of rotations  \cite{gidaris2018unsupervised}. We solely combine FedAvg with AutoEncoder (named \textit{FedAE}), Predicting Rotation (name \textit{FedPR}) and SimCLR (name \textit{FedSimCLR}) respectively and use them as baselines for \textit{FURL}.

\subsubsection{Dataset.} The CIFAR-10/CIFAR100 dataset \cite{krizhevsky2009learning} consists of 60000 32x32 colour images in 10/100 classes, with 6000/600 images per class, and there are 50000 training images and 10000 test images in CIFAR-10 and CIFAR100. The MiniImageNet dataset \cite{vinyals2016matching, deng2009imagenet} is extracted from the ImageNet dataset and consists of 60000 84x84 colour images in 100 classes, we split it into a training dataset with 50000 samples and a test dataset with 10000 samples.
We implement FedCA and the baseline methods on three datasets above in PyTorch \cite{paszke2019pytorch}.

\subsubsection{Federated Setting.} We deploy our experiments under a simulated federated learning environment where we set a centralized node as the server and 5 distributed nodes as clients. The number of local epochs $E$ is 5 and in each round all of the clients obtain global model and execute local training, i.e., the proportion of selected clients $C$ is $1$. For each dataset, we consider two federated settings: IID and Non-IID. Each client randomly samples 10000 images from the entire training dataset in IID setting, while in Non-IID setting, samples are split to clients by class, which means that each client has 10000 samples of 2/20/20 classes of CIFAR10/CIFAR100/MiniImageNet.

\begin{table*}[h]
  \centering
  \fontsize{6.5}{8}\selectfont
  \label{tab:performance_comparison}
    \begin{tabular}{|c|c|c|c|c|c|c|c|c|}
    \hline
    \multirow{2}{*}{Label Fraction}& 
    \multirow{2}{*}{Setting}&
    \multirow{2}{*}{Method}&
    \multicolumn{2}{c|}{CIFAR10}&\multicolumn{2}{c|}{CIFAR100}&\multicolumn{2}{c|}{MiniImageNet}\cr\cline{4-9}
    & &&5-layer CNN&ResNet-50&5-layer CNN&ResNet-50&5-layer CNN&ResNet-50\cr
    \hline
    \multirow{10}{*}{1\%}&
    \multirow{4}{*}{IID}
    &FedAvg (Supervised)&31.84&26.68&9.35&8.09&5.83&5.42\cr
    &&FedAE&35.98&36.86&13.36&14.53&11.71&12.84\cr
    &&FedPR&34.51&36.47&13.15&14.20&11.52&12.34\cr
    &&FedSimCLR&43.95&50.00&22.16&23.01&19.14&19.67\cr
    &&FedCA (ours)&{\bf 45.05}&{\bf 50.67}&{\bf 22.37}&{\bf 23.32}&{\bf 19.20}&{\bf 20.22}\cr\cline{2-9}&
    \multirow{4}{*}{Non-IID}
    &FedAvg (Supervised)&20.99&17.72&6.22&5.37&3.92&3.03\cr
    &&FedAE&23.08&23.43&9.96&9.63&8.45&8.43\cr
    &&FedPR&22.83&23.17&9.83&9.38&8.30&8.58\cr
    &&FedSimCLR&26.08&26.03&14.30&14.02&11.02&10.89\cr
    && FedCA (ours)&{\bf 28.96}&{\bf 28.50}&{\bf 17.02}&{\bf 16.48}&{\bf 13.39}&{\bf 13.03}\cr\hline
    \multirow{10}{*}{10\%}&
    \multirow{4}{*}{IID}
    &FedAvg (Supervised)&50.87&40.44&16.18&14.47&13.46&12.76\cr
    &&FedAE&51.88&53.64&21.77&22.45&21.73&21.96\cr
    &&FedPR&51.38&53.32&21.30&21.21&21.67&21.58\cr
    &&FedSimCLR&59.27&60.67&31.11&31.56&28.45&28.79\cr
    && FedCA (ours)&{\bf 59.91}&{\bf 61.02}&{\bf 31.37}&{\bf 32.09}&{\bf 28.93}&{\bf 29.44}\cr\cline{2-9}&
    \multirow{4}{*}{Non-IID}
    &FedAvg (Supervised)&30.62&21.69&14.90&13.98&11.88&10.13\cr
    &&FedAE&32.07&32.19&18.77&18.98&13.48&13.65\cr
    &&FedPR&31.04&31.78&18.39&18.34&13.30&13.24\cr
    &&FedSimCLR&32.52&33.83&19.91&20.01&15.90&16.03\cr
    && FedCA (ours)&{\bf 35.78}&{\bf 36.28}&{\bf 21.98}&{\bf 22.46}&{\bf 18.67}&{\bf 18.89}\cr\hline
    \end{tabular}
  \caption{Top-1 accuracies (\%) of algorithms for \textit{FURL} on semi-supervised learning}
\label{table-2} 
\end{table*}

\begin{table*}[!t]
  \centering
  \fontsize{6.5}{8}\selectfont
  \label{tab:performance_comparison}
    \begin{tabular}{|c|c|c|c|c|c|c|c|}
    \hline
    \multirow{2}{*}{Setting}&
    \multirow{2}{*}{Method}&
    \multicolumn{2}{c|}{CIFAR100 $\rightarrow$ CIFAR10}&\multicolumn{2}{c|}{MiniImageNet$\rightarrow$CIFAR10}&\multicolumn{2}{c|}{MiniImageNet$\rightarrow$CIFAR100}\cr\cline{3-8}
    & &5-layer CNN&ResNet-50&5-layer CNN&ResNet-50&5-layer CNN&ResNet-50\cr
    \hline
    -&Random init&86.70&93.79&86.60&93.05&58.05&70.52\cr
    \hline
    \multirow{4}{*}{IID}
    &FedAE&87.33&94.23&86.74&94.23&58.82&71.36\cr
    &FedPR&87.22&93.89&87.33&93.55&58.23&70.78\cr
    &FedSimCLR&87.80&94.88&88.03&94.87&59.08&71.85\cr
    &FedCA (ours)&{\bf 88.04}&{\bf 95.03}&87.91&{\bf 94.94}&58.91&{\bf 71.98}\cr\hline
    \multirow{4}{*}{Non-IID}
    &FedAE&87.37&94.35&87.00&94.06&58.56&71.17\cr
    &FedPR&86.97&93.91&86.92&93.55&58.39&70.25\cr
    &FedSimCLR&87.04&94.02&86.81&93.97&58.11&70.91\cr
    & FedCA (ours)&{\bf 87.75}&{\bf 94.69}&{\bf 87.66}&{\bf 94.16}&{\bf 58.93}&{\bf 71.32}\cr
    \hline
    \end{tabular}
  \caption{Top-1 accuracies (\%) of algorithms for \textit{FURL} on transfer learning}
\label{table-3} 
\end{table*}

\subsubsection{Training Details.}
We compare our approach with baseline methods on different encoders including 5-layer CNN \cite{krizhevsky2012imagenet} and ResNet-50 \cite{he2016deep}. The encoder mapping input samples to representations with 2048-dimension and then a multilayer perceptron translate the representations to a vector with 128-dimension used to calculate contrastive loss. Adam is used as optimizer and the initial learning rate is 1e-3 with 1e-6 weight decay. We train models for 100 epochs with a mini-batch size of 128. We set dictionary size $K=1024$, momentum term of temporal ensembling $\alpha =0.5$ and scale factor $\beta=0.01$. 3200 images randomly sampled from STL-10 are used for the alignment module. Data augmentation for contrastive representation learning includes random cropping and resizing, random color distortion, random flipping and Gaussian blurring.

\subsection{Evaluation Protocols and Results}

\subsubsection{Linear Evaluation}
We first study our method by linear classification on fixed encoder to verify the representations learned in \textit{FURL}. We perform \textit{FedCA} and baseline methods to learn representations on CIFAR10, CIFAR100 and MiniImageNet without labels respectively in federated setting. Then we fix the encoder and train a linear classifier with supervision on entire datasets. We train this classifier with Adam as optimizer for 100 epochs and report top-1 classification accuracy on test dataset of CIFAR10, CIFAR100 and MiniImageNet.

As shown in Table \ref{table-1}, federated averaging with contrastive learning works better than other unsupervised approaches. Moreover, our method outperforms all of the baseline methods due to the modules designed for \textit{FURL} as we expect. 

\subsubsection{Semi-Supervised Learning}
In federated scenarios, the private data at clients may be only partly labeled, so we can learn a representation model without supervision and fine-tune it on labeled data. We assume that each client has 1\% and 10\% labeled data respectively. First, we train a representation model in \textit{FURL} setting. Then we finetune it (followed by a MLP consisting of a hidden lay and a ReLU activation function) on labeled data for 100 epochs with Adam as optimizer and learning rate $lr=1e-3$.

Table \ref{table-2} reports the top-1 accuracy of various methods on CIFAR10, CIFAR100 and MiniImageNet. We observe that the accuracy of global model trained by federated supervised learning on limited labeled data is significantly bad, and using the representation model trained in \textit{FURL} as initial model can improve performance more or less. Our method outperforms other approaches, suggesting that federated unsupervised representation learning benefits from designed modules of \textit{FedCA}, especially in Non-IID setting.

\subsubsection{Transfer Learning}
A main goal of \textit{FURL} is to learn a representation model from decentralized and unlabeled data for personalized downstream tasks. To verify if the features learned in \textit{FURL} is transferable, we set the models trained in \textit{FURL} as initial models and then a MLP is used to be trained along with encoder on other datasets. 
The image size of CIFAR (32*32*3) is resized to be the same as MinImageNet (84*84*3) when we fine-tune the model learned from MiniImageNet on CIFAR.
We train it for 100 epochs with Adam as optimizer and set learning rate $lr=1e-3$.

Table \ref{table-3} shows that the model trained by FedCA achieves an excellent performance and outperforms all of the baseline methods in Non-IID setting.

\subsection{Ablation Study}
\begin{figure}[h] 
\centering 
\includegraphics[scale=0.23]{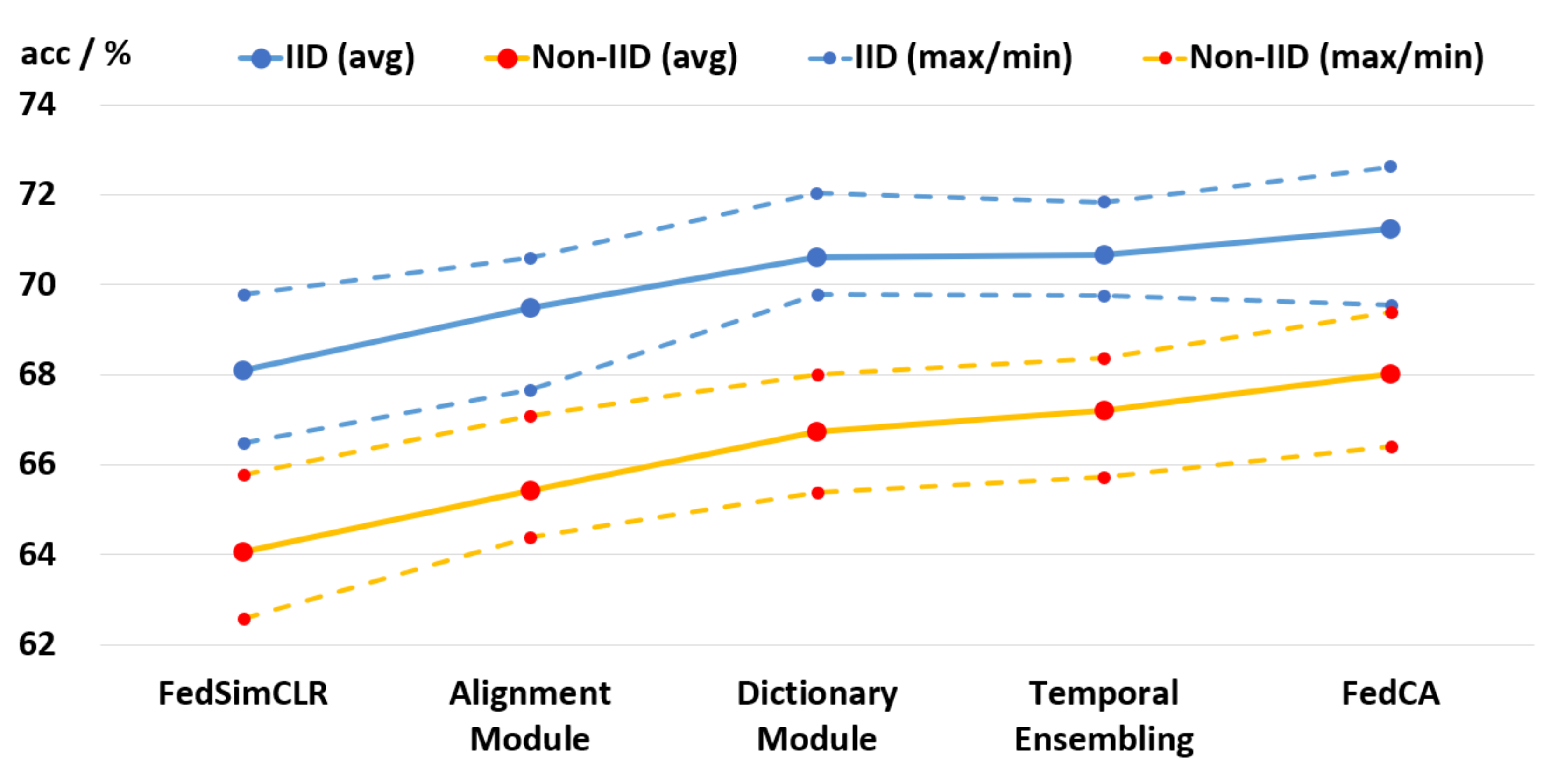} 
\caption{Ablation Study of Modules Designed for \textit{FURL} by linear classification on CIFAR-10 (ResNet-50).} 
\label{fig-5} 
\end{figure}

We perform the ablation study analysis on CIFAR-10 in IID and Non-IID settings to demonstrate the effectiveness of alignment module and dictionary module (with temporal ensembling). We implement (\romannumeral1) FedSimCLR, (\romannumeral2) federated contrastive learning with only alignment module, (\romannumeral3) federated contrastive learning with only dictionary module, (\romannumeral4) federated contrastive learning with only dictionary module based on temporal ensembling, (\romannumeral5) FedCA respectively and then a linear classifier is used to evaluate the performance of the frozen representation model with supervision. Figure \ref{fig-5} shows the results .

We observe that the alignment module improves the performance by ~1.4\% in both IID and Non-IID settings. With the help of dictionary module (without temporal ensembling), there are 2.5\% and 2.7\% increase in accuracy under IID and Non-IID setting respectively. Moreover, we note that the representation model learned in \textit{FURL} benefits from temporal ensembling technique in Non-IID setting than in IID setting, probably because the features learned in IID setting are stable enough so that the temporal ensembling plays a far less important role in IID setting than in Non-IID setting. Fortunately, the model achieves excellent performance when we combine federated constrative learning with alignment module and dictionary module based on temporal ensembling, which suggests that these two modules can work collaboratively and help to tackle the challenges in \textit{FURL}. 

\section{Conclusions}
We formulate a significant and challenging problem \textit{Federated Unsupervised Representation Learning (FURL)} and show two main challenges of this problem: \textit{inconsistency of representation spaces} and \textit{misalignment of representations}. In this paper, we propose a contrastive learning-based federated learning algorithm named FedCA composed of the dictionary module and alignment module to tackle above challenges. Thanks to these two modules, FedCA enables distributed local models to learn consistent and aligned representations while protecting data privacy. Our experiments demonstrate that FedCA outperforms those algorithms that solely combine federated learning with unsupervised approaches and provides a stronger baseline for \textit{FURL}.

In future work, we plan to extend FedCA to cross-modal scenarios where different clients may have data in different modes such as images, videos, texts and audios.

\bibliography{references}
\end{document}